\documentclass{article}
\usepackage{hyperref}       % hyperlinks
\usepackage{booktabs}       % professional-quality tables
\usepackage{microtype}      % microtypography
\usepackage{graphicx}
\usepackage{doi}

\begin{document}
\title{AgRowStitch: A High-fidelity Image Stitching Pipeline for Ground-based Agricultural Images}
\author{Isaac Kazuo Uyehara \\
	University of California, Davis\\
	\texttt{ikuyehara@ucdavis.edu} \\
	%% examples of more authors
	\and
    Heesup Yun\\
    University of California, Davis\\
	\texttt{hspyun@ucdavis.edu} \\
    \and
    Earl Ranario\\
    University of California, Davis\\
	\texttt{ewranario@ucdavis.edu} \\
    \and
    Mason Earles\\
    University of California, Davis\\
	\texttt{jmearles@ucdavis.edu}
}
\date{} 
\maketitle

\section*{Abstract}
    Agricultural imaging often requires individual images to be stitched together into a final mosaic for analysis. However, agricultural images can be particularly challenging to stitch because feature matching across images is difficult due to repeated textures, plants are non-planar, and mosaics built from many images can accumulate errors that cause drift. Although these issues can be mitigated by using georeferenced images or taking images at high altitude, there is no general solution for images taken close to the crop. To address this, we created a user-friendly and open source pipeline for stitching ground-based images of a linear row of crops that does not rely on additional data. First, we use SuperPoint and LightGlue to extract and match features within small batches of images. Then we stitch the images in each batch in series while imposing constraints on the camera movement. After straightening and rescaling each batch mosaic, all batch mosaics are stitched together in series and then straightened into a final mosaic. We tested the pipeline on images collected along 72 m long rows of crops using two different agricultural robots and a camera manually carried over the row. In all three cases, the pipeline produced high-quality mosaics that could be used to georeference real world positions with a mean absolute error of 20 cm. This approach provides accessible leaf-scale stitching to users who need to coarsely georeference positions within a row, but do not have access to accurate positional data or sophisticated imaging systems.

% the * after section prevents numbering
\section{Introduction}
Computer vision techniques and deep learning models are powerful tools for precision agriculture, plant phenotyping, and smart farming (\cite{Mochida2018}, \cite{Tian2020}, \cite{Dhanya2022}, \cite{Ghazal2024}). Agricultural images captured from satellites, drones, robots, and handheld cameras have been used for multiple purposes, including crop monitoring, disease and pest detection, and fruit counting (\cite{Restrepo2024}). While satellite and drone images can easily cover large areas, the resulting image resolution is most suitable for field-scale metrics such as Normalized Difference Vegetation Index (NDVI), plant count, or plant height (\cite{Bauer2019}, \cite{Marques2019}, \cite{Niu2019}). For plant-scale and organ-scale metrics, such as fruit and pest counts, fruit quality, and weed classification, it is more typical to use images taken from ground-based platforms that are close to the crop (\cite{SadeghiTehran2017}, \cite{Halstead2018}, \cite{RicoFernandez2019}, \cite{Madec2023}) or taken in a laboratory setting (\cite{Maharlooei2017}, \cite{Farago2018}, \cite{Wan2018}, \cite{Wang2022}). Automated ground-based imaging of crops is often accomplished through cameras mounted on mobile robots or gantries, which in addition to providing high-resolution images, can also accommodate other sensors placed close to crops and allow for the direct manipulation of plants (\cite{Mueller2017}, \cite{Oliveira2021}, \cite{Vasconcelos2023}).

Although high-resolution images allow for more detailed analyses, there is a tradeoff between image resolution and the number of images needed to capture a region of interest. For example, a few images taken from a drone may cover an entire field, while it may require tens of thousands of images to cover the same area using ground-based imaging. This large collection of images then needs to be georeferenced so that users can analyze how different regions of interest are performing. In addition, for many applications, such as object detection and counting, it is important that the full region of interest can be viewed as a single image. This can be accomplished by creating a single mosaic composed of overlapping images, which allows for objects that appear in multiple images to be only counted once.

While there are many commercial and open source pipelines for stitching drone images, there is no generalized pipeline for creating large mosaics of ground-based images. Generating high-resolution mosaics from ground-based agricultural images is complicated by the large number of images that need to be stitched together, strong parallax when the camera is close to the crop, changes in camera orientation due to uneven terrain, and the loss of distinctive features when the field of view of each image is composed almost entirely of leaves. These issues can be partially mitigated by investing in  ground-based systems with precise GPS and stable camera rigs (\cite{megastitch}), but these solutions may not be accessible or affordable for many users (\cite{Ahmed2024}).

To address the challenges that accompany ground-based agricultural stitching, we created an open access pipeline for stitching high-resolution agricultural images in Python. The pipeline was designed to stitch images taken close to the crop into long mosaics without the need for GPS information or particularly stable camera systems. Instead of stitching an entire field into a single mosaic, our pipeline focuses on the row-scale, where we can assume that sequential images should result in a single straight mosaic. To ensure that our method worked without an expensive imaging system, we tested it with images taken along a row of crops using a camera mounted on a robot with artificial lighting, a camera mounted on a robot with partially controlled lighting, and a camera held by a person walking through the field in ambient light. In all cases, the cameras were less than 1.5 meters from the canopy and the resulting mosaics had mean absolute errors of around 20 cm when measured across a 72 m row.

\section{Related Work}
\label{sec:Related_Works}
Image stitching is generally accomplished by finding features across images, matching those features, projecting each image onto a global surface, and then compositing the final stitched image (\cite{Szeliski2006}, \cite{Brown2007}). Traditional feature detection uses neighborhoods of pixels to develop local descriptions of keypoints using algorithms like SIFT, SURF, and ORB (\cite{SIFT}, \cite{SURF}, \cite{ORB}). Features are then matched across images and a homography or other transform is estimated using a method like RANSAC (\cite{RANSAC}). In regions where multiple images overlap, seam selection and blending can help eliminate artifacts when creating the final stitched image (\cite{seam_carving}, \cite{multiband_blender}). These techniques have been combined into a high-level and powerful stitching pipeline in OpenCV (\cite{OpenCV}).

A major challenge for agricultural stitching is that the accumulation of errors in the stitching process can lead to large-scale drift and distortion in the final mosaic. Drift in the stitching process is particularly problematic for agricultural mosaics because it makes it difficult to accurately georeference the final image for analysis. This is compounded by the fact that images of crops will have many repeating shapes and textures, making accurate keypoint matching more difficult. Moreover, the 3D structure of plants and the fact that plants may move during imaging violates the assumptions of homography transforms, which are only accurate when the scene is planar and static. As a result, stitched images may also contain small-scale errors in overlapping regions that result in objects being blurred, eliminated, or appearing multiple times (ghosting). Stitching errors at this scale make it challenging to reliably count objects in regions of interest.

These general issues have been the topic of previous research, though generally with a focus on aerial images. A notable exception is a stitching pipeline that was tested using a gantry and drone system (\cite{megastitch}). The authors used GPS estimates and ground truth anchor locations to help constrain the stitching process, allowing them to create large mosaics of fields without drift. Researchers improving stitching aerial images have used similar techniques, which take advantage of drone flight path or GPS data to help constrain the stitching process. For example, positional data has been used to estimate the overlapping areas across images. This information can then be used to help constrain feature matching or image projection (\cite{Suzuki2010}, \cite{Zhao2019}, \cite{Cui2021}, \cite{Wang2024}) or remove images with high overlap (\cite{Aktar2020}, \cite{Pham2021}). Stitching images in stages, by first stitching smaller areas together and then stitching those smaller areas into a final mosaic, has also been used to limit drift and distortion (\cite{Aktar2020}, \cite{Liu2022}).

Using customized feature descriptors can also improve feature matching in agricultural settings(\cite{Aktar2020}, \cite{Liu2022}). Newer deep learning techniques may also aid feature detection and matching (\cite{superpoint}, \cite{lightglue}, \cite{Gupta2024}), though these have not been specifically investigated for agricultural stitching. Another strategy to improve stitching is to use non-local warping, which allows for more accurate warping for non-planar scenes (\cite{SVAS}, \cite{Gao2011}, \cite{APAP}, \cite{AANAP} \cite{GSP})). While this technique has been successfully used for drone images (\cite{Pham2021}, \cite{Cui2021}), it is less suitable for high-resolution ground based images, where the depth of the image changes dynamically at the leaf-scale. 

Overall, there are several strategies that have been successfully used to improve agricultural stitching. These include reducing the numbers of images used, restricting pairwise matching, stitching images in batches, using more robust feature descriptors, filtering keypoints based on positional data, and  constraining how images are warped. However, previous methods may not directly translate to ground-based images because of the increased number of images, different field of view, increased parallax and apparent plant movement, and less stable camera orientation. Furthermore, with the exception of improved feature descriptors, these strategies rely on pairing images with positional data that may not be available to all users, especially those operating with a limited budget or in remote areas.

To address these issues, we created an open access and user-friendly pipeline for ground-based images that does not require additional spatial data or particularly stable cameras. We adopt a similar approach as those used for drone imagery, but instead of constraining the stitching process using explicit spatial data, we impose spatial constraints by restricting the pipeline for use on single rows of crops -- effectively trading the generality of full field aerial stitching for the accessibility of high-resolution row stitching.

\subsection{Image Selection}
The core assumption of this method is that images with overlapping fields of view are taken as a camera moves down the center of a single straight row of crops. This allows us to assume that each image will have decreasing overlap with more distant neighboring images and that, while the camera orientation may change and the camera position may move with respect to the center of the row, the camera position will mainly change in one direction across images. Although higher areas of overlap between images should result in more keypoint matches across those images, each area of overlap also introduces potential for stitching errors. Thus, we seek to use the minimum number of images necessary for the final mosaic, while also ensuring that image matches are sufficiently reliable.

\begin{figure*}[h]
    \centering
    \includegraphics[width=0.5 \textwidth]{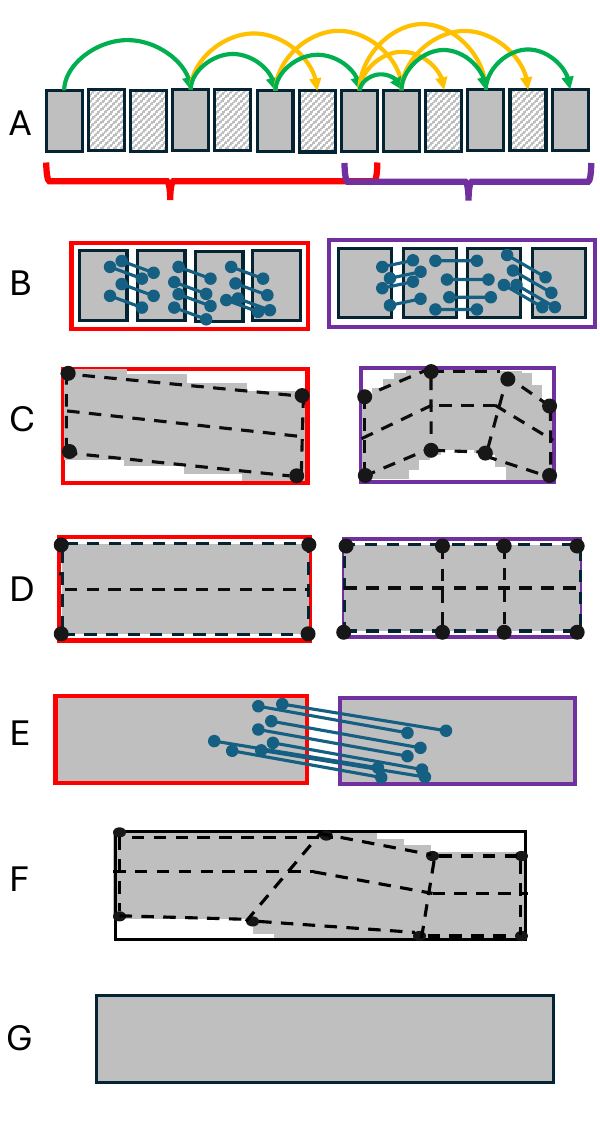}
    \caption{An overview of the stitching pipeline. A) Images are selected in series by attempting to match with more distant images first. When an image matches, the process repeats until a batch of images is complete. The next batch starts with the last image in the previous batch. B) Once the images in a batch have been selected, they are stitched in series. C) Each batch mosaic is sliced into quadrilateral sections and D) those sections are warped into rectangles of constant height. E) The straightened batch mosaics are rescaled and then stitched in series. F) Then the mosaic is straightened again to form G) the final mosaic.}
    \label{fig:stitching_method}
\end{figure*}

To accomplish this, we start with the first image and search within some window of subsequent images to find an image match. The window distance can be set by the user and should depend on the average overlap between subsequent images. The focal image attempts to match with the most distant image in the window and retries with closer images until a match is found (Fig. \ref{fig:stitching_method}A). By starting with the most distant image in the window, we minimize the overlapping area in the mosaic and only use image pairs with higher overlap when necessary. This process is repeated as each new match is found, resulting in a mosaic in which each image is only matched with at most two other images. To determine if two pairs of image are sufficiently well matched, we attempt to match features across the images and estimate a homography and error associated with the homography. If the homography violates constraints imposed by the user, the pair is rejected.

Once a small number of images have been successfully matched in series, this batch is stitched together (Fig. \ref{fig:stitching_method}B-C). Batch size can be set by the user to accommodate different datasets. The next batch starts with the last image of the previous batch as its first image and the process repeats until the final image is matched. At the end of the pipeline, each stitched batch is then stitched to create a final mosaic. This approach is used to improve computational time and reduce drift.

\subsection{Feature Extraction and Matching}

\begin{figure*}[h]
    \centering
    \includegraphics[width=0.8 \textwidth]{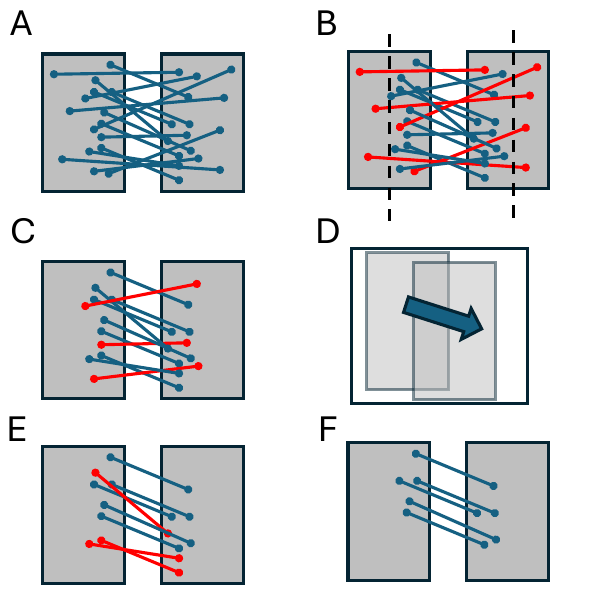}
    \caption{An overview of the image and keypoint match selection algorithm. A) Keypoints are extracted using SuperPoint and then matched using LightGlue. B) Keypoint matches are filtered based on their position relative to the stitching edges of both images. C) A homography is estimated using RANSAC and outlier matches are removed. D) If the camera motion from the homography violates the camera movement assumptions, the match is rejected. E) If the homography was accepted, the remaining matches with the highest reprojection error are removed. F) If the reprojection error of the final matches is smaller than a threshold, the image match and the refined homography are accepted. }
    \label{fig:keypoint_selection}
\end{figure*}

Due to the unreliability of feature matching when the images are composed of many repeating objects with similar shapes and textures (e.g. leaves), we used SuperPoint (\cite{superpoint}) to extract dense features from images and LightGlue (\cite{lightglue}) to match features across images (Fig. \ref{fig:keypoint_selection}A). Since images are ordered and the camera moves down a row, we assume that the overlapping area between a pair of images must be on the side that coincides with forward camera movement for the focal image and the opposite side for a potential matching image. To decrease the chances of a false match between keypoints, we pre-filter matched keypoints by their distance from their stitching edge (Fig. \ref{fig:keypoint_selection}B). Once the matches are filtered based on position, an image pair is rejected if there are insufficient keypoint matches.

If there are sufficient keypoint matches across an image pair, we estimate a homography across the images using RANSAC (\cite{RANSAC}) and remove outliers (Fig. \ref{fig:keypoint_selection}C), with the pair being rejected if the remaining inliers are not above a user-defined threshold. Then we evaluate whether the estimated homography conforms to our assumptions about the camera movement across images (Fig. \ref{fig:keypoint_selection}D). We impose flexible constraints on the homography based on user assumptions about the camera movement during image capture. By default, we assume that the camera moves much more in the direction of the row than in other directions. If this assumption is violated by the estimated homography, the image pair is rejected. If the estimated homography conforms to our movement assumptions, we further filter the matched keypoints by removing matched pairs with the highest reprojection error (Fig. \ref{fig:keypoint_selection}E). This step acts to filter out matched keypoints with low agreement with other points so that matches at differing depths or matches corresponding to parts of the plant that may have moved during imaging can be removed to reduce misalignment. If the final reprojection error passes a threshold, the image pair is accepted as a match and the matched keypoints are stored (Fig. \ref{fig:keypoint_selection}F).

\subsection{Image Warping and Compositing}
Once a batch of images has been matched in series, we fine tune the pairwise homographies with the OpenCV bundle adjustment implementation (\cite{OpenCV}). This bundle adjustment procedure was created for panoramas generated from a stationary, but rotating camera. However, we found that while working on a small batch of images, the OpenCV bundle adjustment produced better results than an affine bundle adjustment, while also avoiding the computational time needed for a bundle adjustment with the full camera motion degrees of freedom. Using the OpenCV implementation also allowed us to use other OpenCV functions for warping and compositing. After bundle adjustment, we adjust the camera orientations using the wave correction function in OpenCV to generate straighter mosaics. Then the images are warped using either a spherical projection or a partial affine transformation, before seam finding and blending to produce the batch mosaic. For these functions, we use the OpenCV warper, color gradient seam finder, and multiband blender functions, respectively.

\subsection{Final Compositing}
After each batch mosaic is composited, they are straightened by warping each mosaic such that the midline of the mosaic becomes a centered horizontal line (Fig. \ref{fig:stitching_method}C-D). To accomplish this, we smooth the edges and midline of the mosaic and identify places where the slope of the midline changes. Then we slice the mosaic into quadrilateral sections such that the midline in each section has a relatively constant slope. Then the mosaic is recomposited by warping each quadrilateral slice into a rectangle of constant height. This  warping does not break the continuity of the adjoining edges of each slice because they are rotated and scaled to the same vertical line.

This technique works to correct for camera movements and misalignment by assuming that each batch mosaic should be rectangular since the camera is generally in the center of the row and the dominant camera movement is translation in the stitching direction. Similar to approaches that use positional data to constrain pairwise homographies, we use the linear nature of a row to force each batch mosaic to be straight. To ensure that the scale across batches is preserved, we rescale each batch mosaic so that their stitching edges are of equal length.

Each straightened and scaled batch mosaic is then stitched end to end with its neighboring batches using SuperPoint (\cite{superpoint}) and LightGlue (\cite{lightglue}) to estimate a 2D translation. Similar to the matching within each batch, only keypoints found at the edges of each batch are used for matching. The warped batches are then composited using the same seam finding and blending procedure as before. The final mosaic is straightened using a similar process as at the batch-scale to create a rectangular mosaic of the full row. This final straightening corrects for any large-scale drift and misalignment from the batch straightening and forces the mosaic to conform to a rectangular mosaic to match the general geometry of a row.

\begin{figure*}[h]
    \centering
    \includegraphics[width=\textwidth]{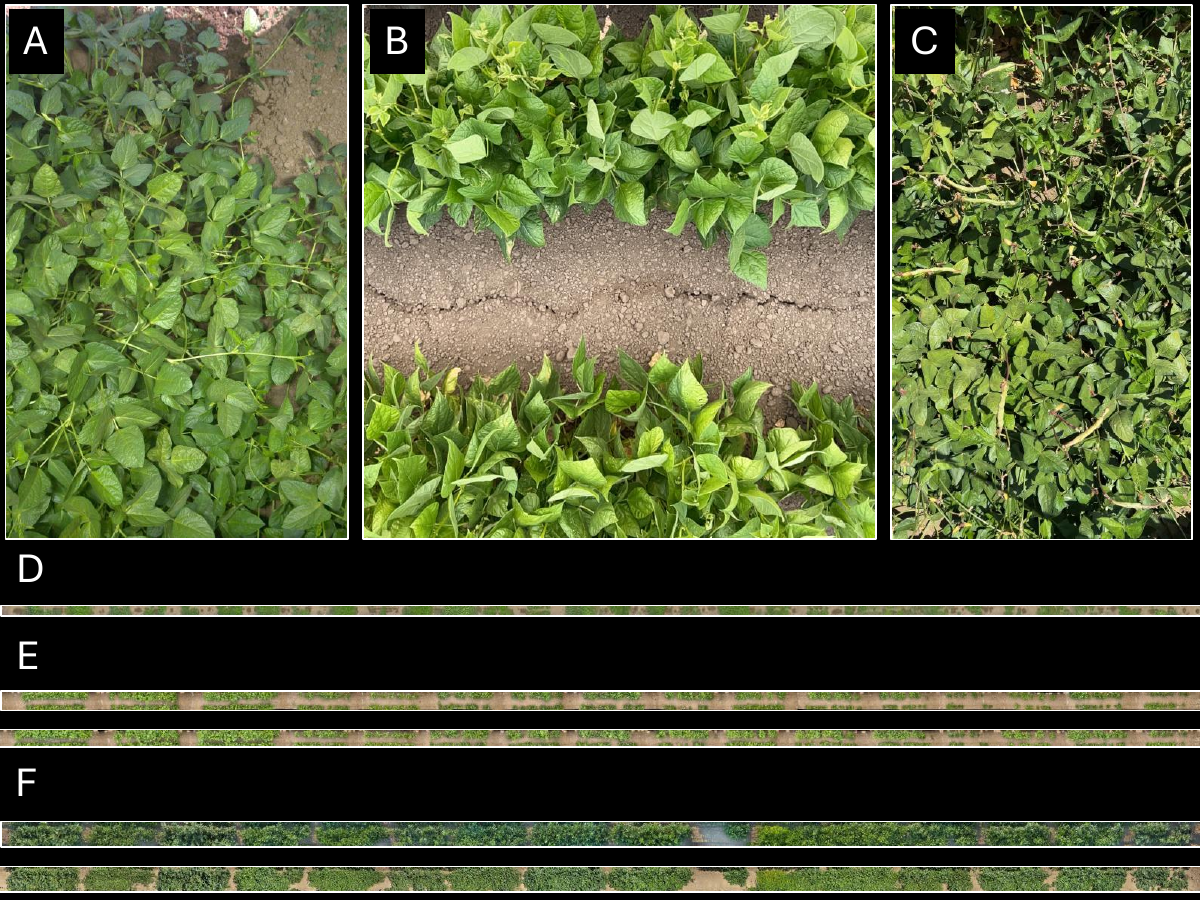}
    \caption{Examples of the input images for the A) T4, B) Amiga, and C) monopod datasets as well as the final mosaics for the D) T4, E) Amiga, and F) monopod datasets. All of the mosaic represents a 72 m row that was scaled to a fixed length.}
    \label{fig:full_mosaics}
\end{figure*}

\section{Datasets}
To test the pipeline we used three datasets that each used a different form of ground-based imaging along a row. All images were taken in fields used for agricultural research. Each row in the field was oriented North and was 72 m long and 1.2 m wide. Each row consisted of 16 plots, with each plot containing a different genotype of a crop. The height of the crops varied across each plot in the row and across datasets, but typically ranged from 10 cm to 1 m in height at the time of imaging. For the following datasets, the mosaics were stitched at full resolution and then resized to a fixed length for analysis.

\subsection{T4}
The first dataset is composed of images taken of a row of cowpeas (\textit{Vigna unguiculata}) on a single day from a 5MP Basler acA2500-20gc camera. The camera was mounted 1.5 m high and oriented downwards on a T4 agricultural robot (Mineral Inc.). The T4 robot creates a partially enclosed environment with artificial lighting as it moves down a row. The camera orientation on the T4 was fixed and there was negligible camera translation outside of the direction of travel. Since the robot was in the camera field of view, the images were cropped to remove the sides of the T4 before analysis, resulting in 1312 x 2048 pixel images. The T4 images are paired with accurate RTK GPS data with 1 cm accuracy. To validate the georeferencing accuracy of our pipeline, we compared the GPS coordinates registered to each image with the pixel coordinates of the center of each of those images in the final mosaic. Mean absolute error (MAE) was calculated based on the pixel deviations from a linear regression of latitude and pixel coordinates. The T4 took 792 images along the row and 507 of the images were used by the pipeline.

\subsection{Amiga}
The second dataset is composed of images taken along a  row of common bean (\textit{Phaseolus vulgaris}) and tepary bean (\textit{Phaseolus acutifolius}) hybrids on two different days. Images were taken using the 12MP camera of an iPhone 13 Pro that was mounted 1.5 m high and oriented downwards on an Amiga agricultural robot (Farm-ng) equipped with light diffusers. Due to uneven terrain, there were minor changes in camera orientation and minor camera translations outside of the direction of travel. As with the T4 data, the images were cropped before analysis so that the Amiga was not in the field of view of the camera, resulting in 1440 x 1500 pixel images. We used small orange stakes that were used to mark plots in the row as ground control points for validation. Of the 14 markers in the row, 11 were visible across both mosaics. MAE was calculated directly on the marker pixel positions across both mosaics. Due to changes in the Amiga driving speed and camera frame rate, different numbers of images were collected on each day. On the first day, 201 images were collected, of which 161 were used by the pipeline. On the second day, 290 images were collected, of which 225 were used by the pipeline.

\subsection{Monopod}
The third dataset is composed of images taken from a row of cowpeas at two different times on the same day. The images were derived from 4K videos shot at 30 frames per second (FPS) taken with an iPhone 13 Pro. Frames from the video were extracted at 10 FPS, generating 2160 x 3840 pixel images. The phone was mounted on a monopod (selfie stick) and carried by a person walking parallel to the row. The monopod was positioned so that the camera was pointed roughly downward and it was roughly centered over the row at around 1.5 m height. However, there were no constraints on the camera movement and orientation while the person walked along the row. Consequently, the height, orientation, and position of the camera all varied noticeablely down the row. In addition, since there was no control on the light environment, the shadows and general lighting changed dramatically across the two imaging times.

We used 14 colored flags placed periodically between the plots as ground control points for validation. MAE was calculated directly on the flag pixel positions across both mosaics. In contrast to the T4 and Amiga, which had fields of view completely within the row, the monopod field of view sometimes includes the furrows between rows. Since walking speed changed between the morning and afternoon imaging, 1647 images were captured in the morning and 1364 were captured in the afternoon. Of these images, 810 and 634, were used by the pipeline, respectively. This dataset was used to test whether the pipeline could be used by users equipped only with a mobile phone camera.

\section{Results}
\begin{figure*}[ht]
    \centering
    \includegraphics[width=\textwidth]{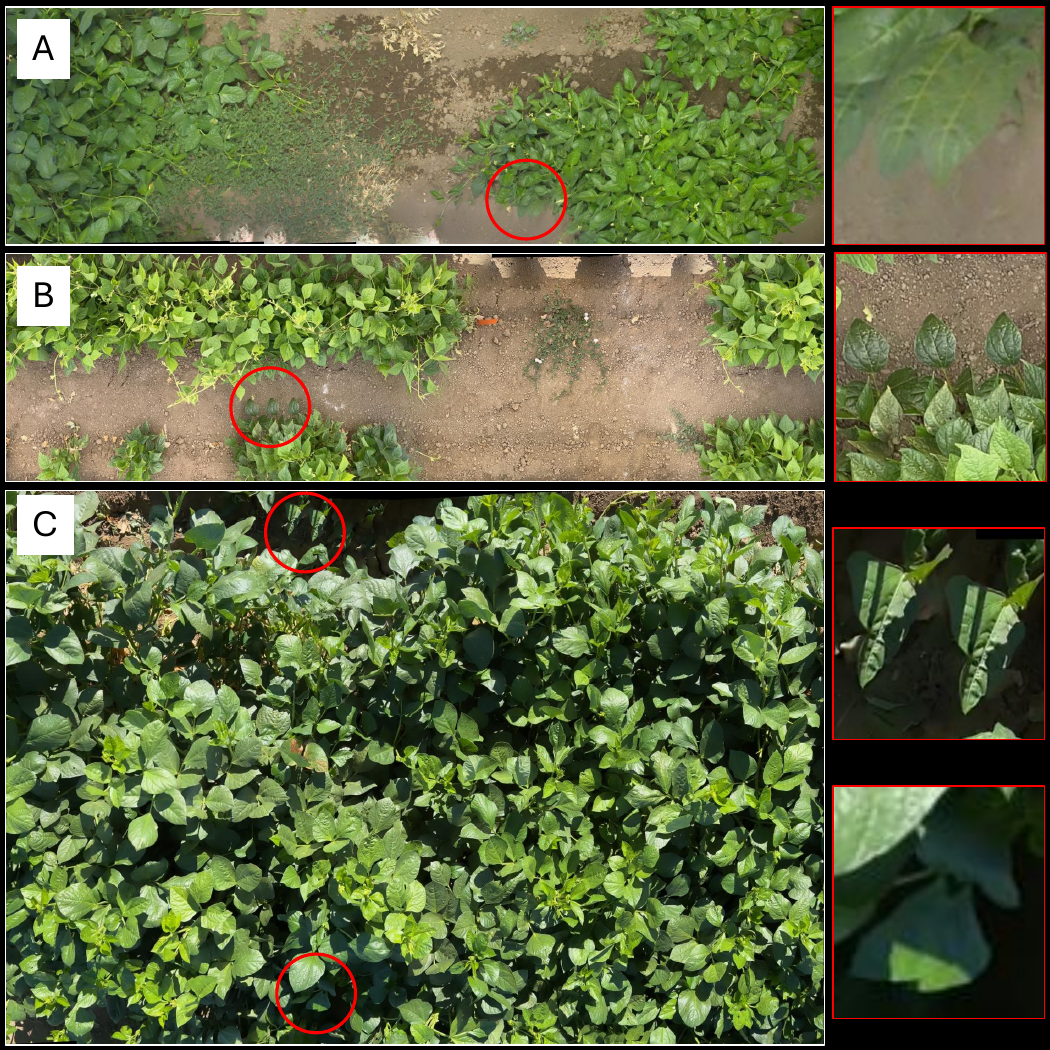}
    \caption{An example of the stitching quality of a single straightened batch mosaic from the A) T4, B) Amiga, and C) monopod datasets. Although there is noticeable blurring and ghosting in some regions of each mosaic, the overall quality is sufficient for object detection. Areas with ghosting are identified in red circles and enlarged to the right. Each batch mosaic is composed of ten images.}
    \label{fig:big_straightened}
\end{figure*}
    
\begin{figure}[h]
    \centering
    \includegraphics[width= \textwidth]{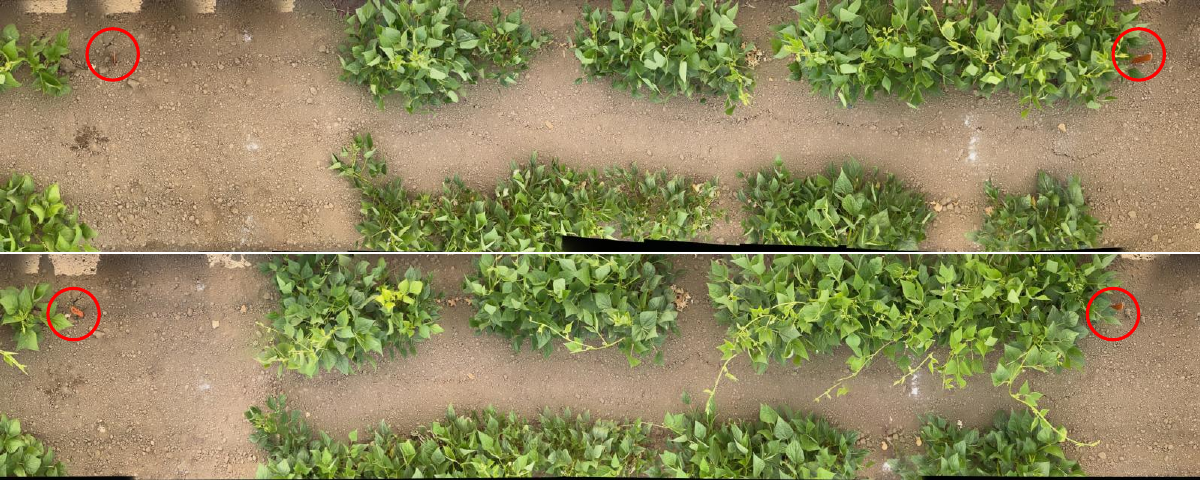}
    \caption{An example of the marker position discrepancies across the Amiga iPhone mosaics.}
    \label{fig:amiga_comparison}
\end{figure}
 
\begin{figure}[h]
    \centering
    \includegraphics[width= \textwidth]{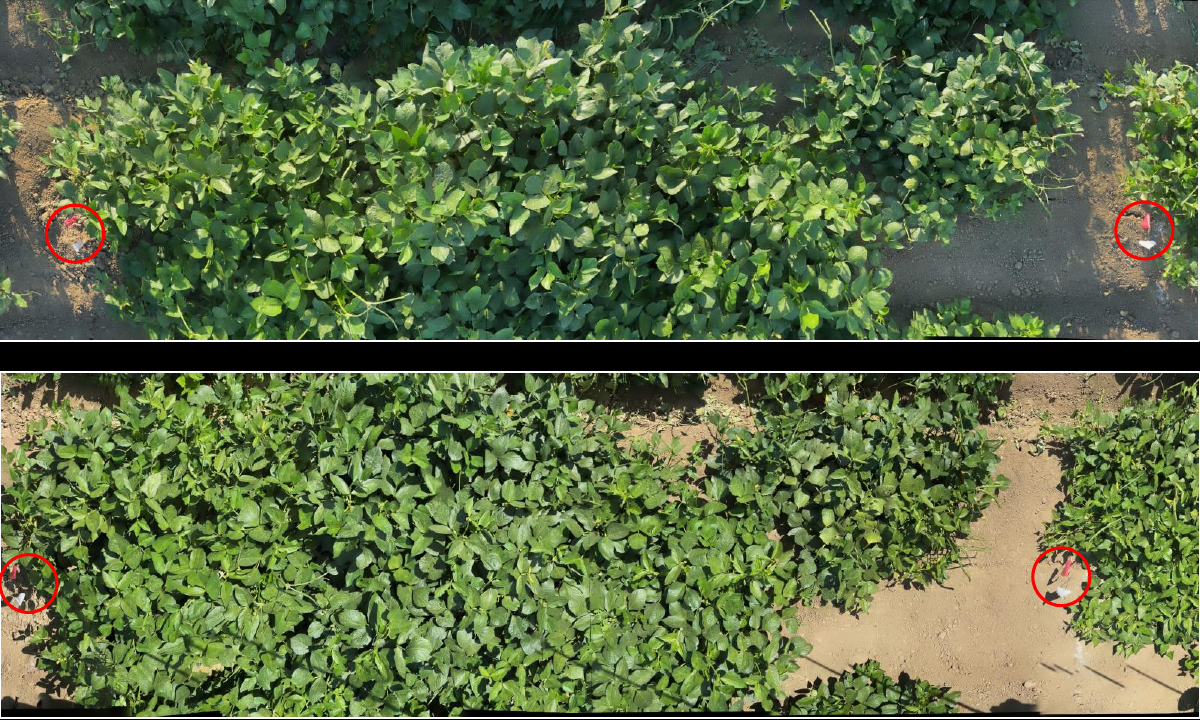}
    \caption{An example of the flag position discrepancies across the monopod iPhone mosaics.}
    \label{fig:kaziga_comparison}
\end{figure}

To be useful for many agricultural applications, the pipeline needs to produce mosaics at the leaf-scale with minimal artifacts. Our pipeline produced high-quality mosaics across all three datasets (Fig. \ref{fig:big_straightened}). Although there are areas in all of the mosaics where there is minor blurring and ghosting, overall, leaves, flowers, and fruits remained intact across seams. The quality of the mosaics was not obviously affected by the stability of the cameras, suggesting that the pipeline can properly handle small changes in camera orientation and position in the non-stitching direction. However, the unstable camera in the monopod dataset may have been offset by the much higher frame rate relative to the other datasets. The reduced field of view in the T4 and Amiga datasets  may have had a negative effect on mosaic quality, especially since the robots sometimes caused plants near the borders to move during imaging.

In addition to leaf-scale stitching, the pipeline was also designed to produce mosaics with high spatial fidelity. The pipeline produces straight mosaics that are free of obvious drift because each mosaic goes through an explicit straightening process. However, misalignments and the straightening process itself can distort the spatial relationships within the final mosaic. To test whether the mosaics could be used directly for georeferencing within the rows, with  pixel coordinates mapping consistently to real-world spatial coordinates, we calculated the MAE of mosaics produced in each dataset. For the T4 dataset, we could directly assess whether pixel coordinates mapped to GPS coordinates. Since we had no access to positional information for the Amiga and monopod datasets, we evaluated the spatial accuracy of the mosaics by calculating the deviation in the pixel coordinates of ground control points when the same row was imaged twice.

\begin{table}[h]
    \centering
    \begin{tabular}{||c|c|c|c||}
    \hline
     Platform & Camera & Control Points & MAE (cm) \\ [0.5ex] 
     \hline\hline
     T4 & 5MP & 507 & 19.8 \\ 
     \hline
     Amiga & 12MP & 11 & 20.4 \\
     \hline
     Monopod & 12MP & 15 & 17.1 \\[1ex] 
    \hline
    \end{tabular}
    \caption{A comparison between the mean absolute error (MAE) across datasets.}
    \label{tab:error}
\end{table}

Across all datasets, the mean absolute error was around 20 cm (Table \ref{tab:error}) across the 72 m rows. Linear regressions of control point coordinates fitted without an intercept in the Amiga and monopod datasets both yielded slopes of 0.999 and $R^2$ of 0.999, suggesting that there was a nearly 1:1 correspondence across ground control points. Thus, even though scale inconsistencies and misalignments propagate along mosaics, resulting in imperfect pixel maps to the real world, our mosaics did not display any consistent biases (Figs. \ref{fig:amiga_comparison} and \ref{fig:kaziga_comparison}). Although errors on the order of 1 cm would allow for more precise georeferencing within each mosaic, an average error of 20 cm over a 72 m row is still sufficient for many applications. For example, in our field, plots within a row were separated by 1.5 m alleys to help distinguish between genotypes. We expect that plot-scale analysis would be generally feasible using our pipeline as long as there is a sufficient buffer between plots.

Given the spatial fidelity across all of the mosaics, our pipeline provides a low-cost alternative to pairing images with precise positional data when errors on the order of 20 cm are tolerable. Our pipeline also makes row-scale stitching accessible to users without access to cameras mounted on agricultural robots. However, the major drawbacks of our method are that it relies on data being organized into individual rows and it does not operate across rows. For users that have positional data paired with images, the pipeline can automatically register these positions in pixel coordinates to provide more reliable georeferencing.

Since errors accumulate during the stitching process, we expect that shorter mosaics will have improved spatial fidelity, as will mosaics generated from datasets with stable high-resolution images taken at a high frame rate. Adjusting the parameters of the pipeline to suit the frame rate and resolution of the imaging platform can help improve the stitching quality or computational time. Users can easily modify these parameters through a yaml file or simply specify a path to the images and use the default settings.

\section{Conclusions}
We created an open access and user-friendly pipeline for stitching ground-based agricultural images that does not rely on additional positional information. To accomplish this, the pipeline only operates on a single linear row of crops at a time. This allows us to impose constraints on the stitching process, resulting in leaf-scale stitching with minimal distortion. Although we found errors on the order of 20 cm when making mosaics of 72 m rows, this accuracy allows for coarse georeferencing within a row. As a result, we believe this pipeline can be particularly useful for agronomists that may not be able to pair their images with reliable positional data or do not have access to an agricultural robot, but can image each row in their field individually. For users with positional data or without the need to georeference within a single row mosaic, the pipeline can be used simply as a stitching pipeline for images taken close to the crop.
%\clearpage

\end{document}